\documentclass[letterpaper]{article} 

\usepackage{aaai2026}

\usepackage{times}  
\usepackage{helvet}  
\usepackage{courier}  
\usepackage[hyphens]{url}  
\usepackage{graphicx} 
\usepackage{natbib}  
\usepackage{caption} 
\usepackage[utf8]{inputenc}
\usepackage[english]{babel}
\usepackage{fontenc}
\usepackage{amsmath}
\usepackage{subcaption}
\usepackage{amsthm}
\usepackage{mdframed}
\usepackage{amsfonts}
\usepackage{tikz}
\usepackage{complexity}
\usepackage{booktabs}
\usepackage{algorithm}
\usepackage{algpseudocode}
\usepackage{blkarray}

\usepackage{breakcites}

\tikzset{
  treenode/.style = {shape=circle, rounded corners,
                     draw, align=center, font=\scriptsize},
  root/.style     = {treenode},
  env/.style      = {treenode},
  leaf/.style     = {shape=rectangle,draw,align=center,font=\scriptsize},
  every node/.style       = {font=\tiny},
  dummy/.style    = {coordinate}
}
\theoremstyle{definition}
\newmdtheoremenv{definition}{Definition}[section]
\newmdtheoremenv{theorem}{Theorem}[section]
\newmdtheoremenv{corollary}{Corollary}[theorem]
\newmdtheoremenv{lemma}[theorem]{Lemma}

\newcommand*{\defeq}{\stackrel{\text{def}}{=}}

\DeclareMathOperator{\load}{load}
\DeclareMathOperator{\length}{len}
\DeclareMathOperator{\emis}{em}

\DeclareMathOperator{\Ad}{Ad}
\usepackage{mathtools}
\DeclarePairedDelimiter\ceil{\lceil}{\rceil}

\addto\captionsenglish{}
\addto{\captionsenglish}{}

\title{Demand Acceptance using Reinforcement Learning for Dynamic Vehicle Routing Problem with Emission Quota
}

\author{
    Farid Najar,
    Dominique Barth,
    Yann Strozecki
}
\affiliations{
    \textsuperscript{\rm 1}DAVID Lab, UVSQ\\

    Versailles, France\\
    first.last@uvsq.fr
}

\begin{document}

\maketitle
\begin{abstract}
   


  This paper introduces and formalizes the Dynamic and Stochastic Vehicle Routing Problem with Emission Quota (DS-QVRP-RR), a novel routing problems that integrates dynamic demand acceptance and routing with a global emission constraint. A key contribution is a two-layer optimization framework designed to facilitate anticipatory rejections of demands and generation of new routes. To solve this, we develop hybrid algorithms that combine reinforcement learning with combinatorial optimization techniques. We present a comprehensive computational study that compares our approach against traditional methods. Our findings demonstrate the relevance of our approach for different types of inputs, even when the horizon of the problem is uncertain.
\end{abstract}


\medskip

\section{Introduction}

Due to climate change, increasingly strict regulations are being introduced to limit greenhouse gas emissions~\cite{MT17} and other pollutants. Consequently, transport companies must comply with emission constraint within specific areas, sometimes expressed as hard quotas. To meet these constraints, some deliveries need to be omitted. However, in many supply chain contexts, delivery requests are not fully known in advance and are revealed dynamically over time.


In this work, we address a \textbf{Capacitated Vehicle Routing Problem (CVRP)} with heterogeneous vehicles, a hard emission constraint, and online demand acceptance. We model a transporter which receives dynamic requests of clients for a certain day or week and must decide immediately whether to accept each new demand. The acceptation and rejection are done online during the planning phase, before the beginning of the tour. The objective of the transporter is to maximize the number of requests served, while respecting the capacity constraints and  the emission limitation. The aim of this work is to solve this problem in a setting where the distribution of the demands is known, a reasonable assumption since this distribution can be easily learned through historical data.

There exist multiple versions of VRP where dynamism or stochasticity arises in some part of the problem. We study here the most classical case where the requests are revealed in an online fashion and are drawn from a known distribution, hence the problem is a \textbf{Dynamic and Stochastic VRP with Random Requests (DS-VRP-RR)}~\cite{zhang_dynamic_2023}~\footnote{Other names exist for this problem like (Dynamic and) Stochastic Vehicle Routing Problem with Random or Stochastic Customers (DS-VRP-SC) among others, but all designate a similar or equivalent problem.}
The main specificities of our model are the hard quota on the emission and the necessity to accept and reject requests in an online fashion before the tour has even begun. We call our problem \textbf{Dynamic and Stochastic VRP with emission Quota and Random Requests (DS-QVRP-RR)} and for simplicity, we denote it by DQVRP in the rest of this work.

The problem DQVRP we introduce can be related to the \textbf{Team Orienteering Problem (TOP)}~\cite{chao1996team}, and its capacitated version~\cite{archetti2009capacitated}. In TOP, the objective is to find a tour which maximizes the total rewards collected, while each vehicle has to respect a constraint on the total cost of its route (typically a maximum time). In our problem, the constraint is global and is imposed on the total emissions of the vehicles, which are heterogeneous.

\subsection{Related Works}

The static version of the demand-selection problem under an emission constraint was studied in~\cite{najardemand}, where the authors proposed a hybrid RL/OR approach and showed that classical OR methods outperform RL in this setting. More broadly, the static DQVRP is related to the Team Orienteering Problem (TOP), which has been addressed through both OR and ML/RL approaches~\cite{lahyani:hal-01663624,kant2022orienteering,xiao2022benchmark}. Recent works have shown that transformer-based models can match or surpass state-of-the-art OR methods on the TOP~\cite{sankaran2022gamma,fuertes2025top}.

Dynamic VRPs with random requests (DVRP-RR) have been extensively studied due to their relevance in modern logistics. A recent survey~\cite{zhang_dynamic_2023} distinguishes three categories: service requests, pickup-and-delivery, and delivery requests. Our problem belongs to the delivery-request category, where all deliveries originate from a known hub. However, unlike most DVRP variants, the tour in our setting is executed after all requests arrive, while accept/reject decisions must be made dynamically. Classical DVRP models rarely allow anticipatory rejection; when rejections occur, they typically happen post hoc, once service becomes infeasible. Anticipatory methods generally rely on costly simulation-based rollouts, such as the Multiple Scenario Approach (MSA)~\cite{bent_scenario-based_2004} where many scenarios are generated, and the decision is taken by a consensus choice on the solutions to these scenarios. Similarly, \cite{ulmer2019anticipation} compares reactive re-optimization (RHR) using MIP with anticipatory rollout (RA), showing that RA performs better in fully dynamic settings by incorporating future rewards, while RHR dominates RA when there are only a few dynamic requests.

Dynamic variants of the TOP have also been considered. In~\cite{kirac2025dynamic}, the authors introduce a dynamic TOP where new requests arrive during the tour, in contrast to our setting where the tour starts only after all requests have been revealed. The authors propose the Multiple Planning Approach (MPA), which maintains and updates a pool of alternative routing plans as new requests are revealed dynamically.

Many DVRP-RR formulations rely on Markov decision processes (MDPs)~\cite{psaraftis_dynamic_2016,zhang_dynamic_2023}. For specific cases such as a single uncapacitated vehicle with a time horizon, analytical results on optimal policies exist~\cite{thomas2007}. More recently, \cite{zhang_solving_2023} proposed potential-based lookahead policies using knapsack-based approximations to predict the expected reward-to-go, without performing extensive simulations.

There is also growing interest in using RL for both assignment and routing decisions. For VRPDSR with homogeneous vehicles, \cite{bono2020solving} models each vehicle as an agent in a multi-agent MDP and updates routes dynamically as new requests appear, although request rejection is not permitted. Closer to our work, \cite{chen2022deep} (SDDP) and \cite{kullman2022dynamic} (DPDP) apply DQN-based policies for accept/reject decisions in settings involving EV fleets or heterogeneous fleets with drones. In both cases, RL methods with carefully designed features outperform heuristic using a threshold and accept-all baselines, highlighting the potential of RL for anticipatory decision-making and the importance of the choice of features.

\subsection{Our Contributions}

We introduce the DS-QVRP-RR (DQVRP), a dynamic variant of the QVRP that incorporates real-world constraints such as emission quotas and stochastic demand arrivals. We propose a two-layer solution framework combining a reinforcement-learning module for demand acceptance and vehicle assignment with lightweight routing heuristics. The RL component leverages engineered state features to enable strategic acceptance, rejection, and assignment decisions.

We benchmark our approach against the First-Arrived-First-Served (FAFS) heuristic which accepts incoming demands sequentially and recalculates routes accordingly, and the Multiple Scenario Approach (MSA), a representative sampling-based method. Experimental results show substantial improvements in service level, achieved with both computational efficiency and robustness across diverse operational conditions. Our findings provide actionable insights for fleet operators seeking to optimize dynamic routing under environmental constraints and contribute to advancing research on dynamic VRPs and dynamic TOPs.

\section{Problem Setting}
\label{sec:problem}

We model a transporter that wants to serve demands with its fleet of vehicles under an emission constraint. 
We denote the set of integers $\{1,\dots,K\}$ by $[K]$, and we represent the potential destinations by integers in $[K]$, while the hub vertex is represented by $0$. The distance matrix between all pairs of destinations is $D$, thus $D_{i,j}$ ($=D_{j,i}$) is the distance to travel from $i$ to $j$.

A \textbf{route} $R$ is an ordered list of destinations in $[K]$, without repetition, which begins and finishes with the hub vertex $0$: $(d_0 = 0,d_1,\dots,d_k, d_{k+1} = 0)$. The length of a route $R$ is the sum of the distances between consecutive destinations in $R$: $\length(R) \defeq \sum_{i=0}^{k} D_{d_i,d_{i+1}}$. We denote by $\mathcal{D}(R)$ the set of destinations reached on the route $R$, that is, all $d_i$ for $1 \leq i \leq k$. We let $\load(R) = k$,
the number of destinations served along the route.

We let $\mathcal{V}$ be the set of \textbf{vehicles} of the transporter, all of the same capacity $C$.
Each vehicle $v \in \mathcal{V}$ has an emission factor $Ef_v$ which models the type of the vehicle and its engine technology. In this work, we have chosen to make emission depend linearly on the distance, i.e. on the length of the route. Thus, the emission of a vehicle $v$ on a route $R$ is defined as $\emis_v(R) = Ef_v \times \length(R)$.

A \textbf{routing} $\mathbf R$ is a sequence $\mathbf R = (R_v)_{v \in \mathcal{V}}$, where the $R_v$ are routes such that no destination appears in two routes. We assume that a vehicle brings the same quantity of goods to each destination, and we ask that each vehicle transports less than its capacity: $\load(R_v) \leq C$.  We define the emission of a routing $\mathbf R$ as the sum of the emissions of its vehicles : $\emis(\mathbf R) \defeq \sum_{v \in \mathcal{V}} \emis_v(R_v)$. In our settings, the transporter must respect strong emission regulations set by governing bodies, represented by an emission quota denoted by $Q$. A routing $\mathbf R$ is \textbf{feasible} only if its emissions are less than the quota: $\emis(\mathbf R) \leq Q$. 

We set some horizon $H\in\mathbb{N}^*$ and we define a \textbf{scenario} as $s = (d_1, \dots, d_{H})$, where each $d_t \in [K]$ is a destination, which represents a demand from a client.  We are given a probability distribution $P$ over the destinations in $[K]$ and from that we define a distribution $\mathcal{P}_{H,P}$ on the set of scenarios of horizon $H$: The destinations are successively sampled without replacement according to $P$.

For a scenario $s$, we let $\mathcal{D}_s$ be the set of all destinations of $s$. We let the set of destinations served by a routing  $\mathbf R$ be $\mathcal{D}(\mathbf R) = \cup_{v \in \mathcal{V}} \mathcal{D}(R_v)$. The routing $\mathbf R$ is a solution for $s$ if $\mathcal{D}(\mathbf R) \subseteq \mathcal{D}_s$. Typically, the constraint on emissions cannot be satisfied while serving all destinations in $\mathcal{D}_s$: the transporter has to omit some destinations. The objective is then to maximize the number of destination served by the routing, that we call the accepted demands:   $\Ad(\mathbf{R}) \defeq \sum_{R \in \mathbf R} \load(R)$.

A share of the demands is known in advance and represents regular clients that are guaranteed to be served. The remaining clients reveal their demand in an online fashion and can be accepted or rejected. Their share is defined by $\delta$ the \textbf{Degree of Dynamism (DoD)}, that is the proportion of unknown demands.
In our model, at each timestep $t$, we discover the demand $d_t$, and some policy decides whether $d_t$ must be served. More formally, we let $I = (\mathcal{V},D,P,Q,H,\delta)$ be an instance of our problem. A history of size $t$ is a sequence $(d_1,a_1),\dots,(d_t,a_t)$ of pairs of destinations $d_i$ and action $a_i \in \{0,1\}$ where $a_i = 0$ means that $d_i$ is rejected. A \textbf{policy} $\pi_I$ for the instance $I$ is a function, with arguments a history $h$ of size less than $H$ and a destination $d$, which returns $0$ if $\pi_I$ rejects $d$ and $1$ if it commits to serve it.

From a scenario $s = (d_1,\dots,d_H)$, a policy $\pi_I$ defines an \textbf{omission assignment}, that is a function from $[H]$ to $\{0,1\}$ as follows. We let $\alpha(1) = \pi_I(h_{0},d_1)$, where $h_0$ is the history of size $0$. Then, for all $t \leq H$, $\alpha(t) = \pi_I(((d_1,\alpha(1)),\dots(d_{t-1},\alpha(t-1))),d_t)$. We denote the omission assignment $\alpha$ produced by $\pi_I$ on the scenario $s$ by $\pi_I(s)$.
We say that a policy respects the degree of dynamism $\delta$, if it accepts all demands known in advance, that is for all $t \leq \ceil*{(1-\delta)H}$, $\pi_I(s)(t) = 1$. We say that a routing $\mathbf R$ for a scenario $s = (d_1,\dots,d_H)$ is compatible with the omission assignment $a$ if $\mathcal{D}(\mathbf R) = \{ d_i \mid i \in a^{-1}(1)\}$.
We say that a policy $\pi_I$ is \textbf{admissible}, if for all $s$, it respects the degree of dynamism and there exist an admissible routing compatible with $\pi_I(s)$.

Note that for a scenario $s$, all routings compatible with an omission assignment $a$ have the same number of accepted demands, that we denote by $\Ad(a,s)$.  We can now formally define the problem we tackle, the Dynamic and Stochastic Vehicle Routing Problem with Emission Quota and Random Requests (DS-QVRP-RR or DQVRP for short):  given an instance $I$, find a policy $\pi_I$ which maximizes the average number of accepted demands $\mathbb{E}_{s\sim \mathcal{P}_{H,P}}[\Ad(\pi_I(s),s)]$.

\subsection{Algorithmic framework}\label{sect:two layers}

\paragraph{Two layers decomposition}

In this work, we decompose the hard problem of solving DQVRP into two subproblems or layers.
The first layer, the \textbf{Assignment Layer} must decide which demands to accept, while the second layer, the 
\textbf{Routing Layer}, must find a routing respecting the choices of the assignment layer with the smallest possible emission. 
This allows to decompose the algorithms solving DQVRP into two different and mostly independent parts, where one can be changed without changing the other.

A \textbf{vehicle assignment} $\alpha$ for a scenario $s=(d_1,\dots,d_H)$ is a mapping from $[H]$ to $\{ 0 \}\cup \mathcal{V}$, it is a refined version of an omission assignment, where each demand is mapped to a vehicle or to $0$, which means it is rejected. We say that a routing $\mathbf R$ respects the vehicle assignment $\alpha$ if $\mathcal{D}(R_v) = \{ d_i \mid i \in \alpha^{-1}(v)\}$. In the following, we give two kinds of algorithms for the Assignment Layer, which computes either an omission assignment or a vehicle assignment dynamically. 
  
The routing layer is abstracted by a function $\rho(I,\alpha,s)$ returning a routing for the scenario $s$ where $\alpha$ is the omission or vehicle assignment given by the Assignment Layer and $I$ the instance of the problem. The routing $\rho(I,\alpha,s)$ must respect $\alpha$, the capacity constraint and the emission constraint given in $I$, which is achieved by trying to minimize the emissions. Solving the routing layer given a vehicle assignment corresponds to solving one TSP for each vehicle, while solving it given an omission assignment corresponds to solving a CVRP on the accepted demands. Hence, choosing to compute a vehicle or an emission assignment allows to differently balance the work in the two layers: more work in the assignment layer for the vehicle assignment and more work in the routing layer for the omission assignment.

\paragraph{Offline and online phases}

 In DQVRP, we are given a degree of dynamism $\delta$, with the constraint that all destinations 
 $d_1,\dots,d_{\lceil \delta H \rceil}$ must be accepted. Hence, all algorithms have first an \textbf{offline phase} where it computes the best possible routing for these destinations and if it does not satisfy the emission constraint,  the full algorithm fails. Note that this part is done in one step and involves only the routing layer, and may thus use a more expensive algorithm to solve it.
 
Then, the algorithm enters its \textbf{dynamic phase}, where the Assignment Layer has computed an assignment for the destinations  $d_1,\dots,d_i$ for $i < H$ and receives a new destination $d_{i+1}$ and must decide whether to accept it.  If the demand is accepted, the routing layer computes a new routing respecting the new assignment and the emission quota, and if it fails to do so, the demand is eventually rejected. To help the routing layer, we give it access to the routing it has computed in the previous step.

\section{Methods}

In this section, we first present the methods used to compute routes in the routing layer. These procedures are intentionally kept simple and computationally efficient, ensuring fast experimentation and enabling effective training of the assignment-layer algorithms, which constitute the core contribution of this work.

We then describe two classical baseline methods for the assignment layer: the \textit{First Arrived, First Served} (\texttt{FAFS}) heuristic and the \textit{Multi-Sampling Approach} (\texttt{MSA}). These baselines provide reference performance levels for evaluation.

Finally, we introduce our Reinforcement Learning (\texttt{RL} assignment policy. The goal is to design an efficient online algorithm that consistently outperforms the baselines and approaches the performance of an offline oracle with full knowledge of all future demands.

\subsection{Routing Method}
\label{sec:routing}

In the offline phase, the routing layer must satisfy all given demands. It first compute an initial solution by including all the accepted demands using the \textbf{Nearest Neighbor algorithm}. This algorithm greedily builds a route for each vehicle, by selecting the cheapest insertion in the routes of the vehicles which still have remaining capacity.

Then, the routing layer tries to minimize the emission of the initial routing using \textit{Simulated Annealing}  (SA)~\cite{metaheuristics}. More specifically, the SA generates neighbors using as neighborhood operators \emph{insertion} (inserting a randomly chosen destination into a random position), \emph{swap} (changing the positions of two randomly chosen destinations) and \emph{2-opt} (reversing a segment of the permutation). Each operator is used with respective probabilities 30\%, 60\% and 10\%. The temperature decreases using an exponential cooling\footnote{Hyper-parameters are available in the Appendix}.

During the dynamic phase, the routing layer does a different computation whether it is given an omission assignment or a vehicle assignment to respect. In both cases, it has access to the previous routing containing all accepted destinations but the new one.  

When the routing layer is given an omission assignment, it acts as a continuation of the algorithm for the offline part: The new destination is inserted in the current routing at the position which minimizes the additional emission and without violating the capacity constraints. Then, SA is applied to improve the routing. 

When the routing layer is given a vehicle assignment, there is an additional constraint that each destination must be served by a specific vehicle. Hence, the new destination is inserted in the vehicle given by the vehicle assignment, at the position which minimizes the emission. Then, SA is applied only to the route where the insertion has been made, to avoid exchanging destinations between routes.

\subsection{DQVRP Model}\label{sec:env}

From now on, we abstract the routing layer as a black-box entity. This entity is completed with additional elements, such as observations and rewards, to become a MDP \emph{environment}. And we call \textbf{agent}, any decision making algorithm that computes a policy for this MDP and which thus solves $\textsc{DQVRP}$.

\subsubsection{MDP Formulation}\label{sec:MDP}

We adopt a standard MDP formulation commonly used in reinforcement learning. In this framework, the environment starts in an initial state and interacts with the agent until a terminal state is reached. The sequence of interactions from the initial to the terminal state defines an \textit{episode}.

For our problem, the MDP is defined from an instance 
\( I = (\mathcal{V}, D, P, Q, H, \delta) \) of the \textsc{DQVRP}.  
A state at time \( t < H \) encodes: (i) the current time step, (ii) the first \( t-1 \) demands and the assignment decisions made for them, (iii) the current routing, and (iv) the newly revealed demand \( d_t \).  
The action space is \( \mathcal{A} = \{0\} \cup \mathcal{V} \), where \( 0 \) denotes rejection and \( v \in \mathcal{V} \) denotes assigning \( d_t \) to vehicle \( v \).

After action \( a_t \), the environment transitions to time \( t+1 \).  
The routing layer updates the routes accordingly and may invalidate an acceptance decision if \( d_t \) cannot be feasibly inserted, in which case the demand is effectively rejected. The new demand $d_{t+1}$ is drawn according to the probability distribution $P$. 
The initial state corresponds to the offline phase ($t = \delta H$), while terminal states occur when the horizon is reached (\( t = H \)).

The reward function is defined as \( r_t = 1 \) if \( d_t \) is accepted and successfully integrated into the routing, and \( r_t = 0 \) otherwise. The cumulative reward over the episode therefore corresponds to the total number of accepted demands under the policy.

In our RL framework, providing the full environment state to the agent is infeasible. The observation must therefore be engineered to be both informative and compact: an overly rich representation leads to the \textit{curse of dimensionality}, while an under-specified one prevents effective decision-making. We thus construct a partial observation that selectively retains only the information relevant for routing decisions.

\paragraph{Distance to assigned destinations.}

To characterize the position of the incoming demand $d_t$ relative to the current routing $\mathbf R^t = (R_{v_1}^t,\dots,R_{v_m}^t)$, we measure its proximity to each vehicle's already assigned destinations. Since insertion into a route is performed near its closest points, only the nearest neighbors are required. Let $\text{NN}_D(k, d, S)$ denote the $k$ nearest neighbors of $d$ in the set $S$ under distance metric $D$. For each vehicle $v$ with route $R_v$, we define
\[
    \mathrm{DAD}(k_\mathrm{FD}, d_t, R_v)
    =
    \frac{E f_v}{k_\mathrm{AD}}
    \sum_{d \in \text{NN}_D(k_\mathrm{AD}, d_t, \mathcal D(R_v^t))}
        D_{d,d_t}
\]
 This feature reflects the proximity of $d_t$ with destinations assigned to the vehicle $v$ and thus the potential marginal emission cost of assigning this destination to $v$.

\paragraph{Distance to potential future demands.}
We also characterize $d_t$ with respect to destinations that may appear later in the scenario. Let $FD(s)$ denote the set of future destinations, that is, which are not in the scenario $s$. Among the $k_\mathrm{FD}$ nearest future destinations, we weight distances by their occurrence probability $p_i$ and take the median to reduce sensitivity to outliers:
$$
    \mathrm{DFD}(k_\mathrm{FD}, d_t)
    =
    \operatorname*{median}_{d \in \text{NN}_D(k_{\mathrm{FD}}, d_t, FD(s))} \bigl( p_d\, D_{d,d_t} \bigr)
$$
This signal helps identify demands located in regions with high likelihood of future arrivals, allowing the agent to exploit spatial demand structure.

\paragraph{Observation vector.}

At time $t$, given routing $\mathbf R^t = (R_1,\dots,R_m)$, the observation vector $x_t$ is
\begin{align}\label{eq:observation}
  \newcommand\col[1]{\parbox{1cm}{\centering\scriptsize#1}}
  \newcommand\row[1]{\text{\scriptsize#1}}
  x_t =
  \begin{blockarray}{cccc}
  \begin{block}{[ccc]c}
           &H - t && \\
           &C - \load(R_1^t) && \\
           &\vdots && \\
           &C - \load(R_m^t) && \\
           &Q - \emis(\mathbf R^t) && \\
           &DAD(k_{\text{AD}}, d_t,  R^t_1)&&\\
           &\vdots && \\
           &DAD(k_{\text{AD}}, d_t, R^t_M)&&\\
           &DFD(k_{FD},d_t) && \\
  \end{block}
  \end{blockarray}
\end{align}

This observation captures three essential aspects for solving \textsc{DQVRP}:  
(i) the number of remaining demands,  
(ii) the operational state of the fleet (residual capacities and quota), and  
(iii) the position of the current demand with regard to past and potential future demands.

\subsection{Assignment Methods}\label{sec:assignment}
To evaluate our methods, we compare it with two classical methods that are used either in practice or by academics. The policy First Arrived, First Served, \texttt{FAFS}, is the by default strategy of the transporters as they prefer not to take the risk of rejecting a customer. 
The \texttt{MSA} method is the most classical choice when designing an algorithm dealing with uncertainty. Furthermore, we consider an \texttt{Offline} algorithm to serve as an upper bound for the dynamic methods.

\subsubsection{Baselines}\label{sec:baselines}

\paragraph{Offline Agent}\label{alg:offline}
The \texttt{Offline} agent does not solve the dynamic problem DQVRP, but a static version of it, where it has access to the whole scenario before deciding which destinations to serve. Since it solves a relaxed problem, without uncertainties, it should give us an upper bound on the quality of the solution found by the other algorithms. 

The \texttt{Offline} agent must compute an omission assignment $\alpha$ for the whole scenario $(d_1,\dots,d_H)$. The agent cannot use the routing layer for that task, since it must reject some destination, but we show how to adapt it. To compute the omission assignment, instead of considering routings, we consider all permutations of the destinations of the scenario. From a permutation, we use a greedy split algorithm to reconstruct the \emph{associated routing}: it affects the destinations to a vehicle in order of the permutation, until its capacity is met or the emission quota has been exceeded. The vehicles are considered in increasing order of emission factor. Because of the constraint on capacity and emission, the last destinations in the permutation may not be in the associated routing, which means they are rejected.  

We want to find the permutation such that its associated routing maximizes the number of accepted demands. To do that, we perform simulated annealing on permutations, using the same neighborhood as in the routing layer (insertion, swap, 2-opt). However, many solutions have the same number of accepted demands, hence to better guide the SA, we define the refined objective function $h(\mathbf{R}) = L\times \Ad(\mathbf{R}) + \emis(\mathbf{R})$, where $L$ is twice the largest emission when going from a destination to another. By using $h(\mathbf{R})$ instead of $\Ad(\mathbf{R})$, we favor among two solutions with the same number of accepted demands the one with less emissions.

\paragraph{First Arrived, First Served}\label{fafs}
The \texttt{FAFS} agent computes an omission assignment without taking into account the future. It accepts all incoming demands, and asks the routing layer to integrate the new demand into the routes. Thus, the agent only rejects a demand, when the routing layer fails to satisfy the constraints.

This strategy is widely adopted in real-world operations since transporters typically either can serve all demands, or follow a simple first-come-first-served policy to process requests in their arrival order.

\paragraph{Multi Sampling Approach}\label{msa}
The \texttt{MSA} agent samples a number of scenarios, compute good routings for these scenarios, and decides from these routings whether to accept or not the current demand. 
Let us precisely describe how the \texttt{MSA} agent deals with a new demand $d_t$, after having computed 
an omission assignment $\alpha$ for $d_1, \dots, d_{t-1}$. The agent builds a set of scenarios $S= \{s_1, \dots, s_n\}$, all of which extends $d_1, \dots, d_{t}$ to a scenario of size $H$. The scenarios are drawn over all possible scenarios (without replacement), following the distribution $\mathcal{P}_{H,P}$ given in the instance. In this work, the \texttt{MSA} agent performs $n=101$ simulations.

Then, the \texttt{MSA} agent must solve each scenario of $S$, which is an offline version of our problem, with the additional
constraint that the solutions must respect the omission assignment $\alpha$ on the first $t-1$ destinations. We are given a scenario $s\in S$, where $s=(d_1,\dots,d_{t},\dots,d_H)$ and the routing $\mathbf{R}$ produced on $d_1, \dots, d_{t-1}$ and satisfying $\alpha$. We add destinations in $\{d_{t},\dots,d_H\}$ to the routing $\mathbf{R}$ using the cheapest insertion heuristic (insert the destination which incurs the smallest emission increase and which respects the capacity). The algorithm stops when all destinations are inserted or the quota $Q$ is violated.

After computing the solutions for each scenario, the \texttt{MSA} agent conducts a voting process to determine whether $d_t$ should be accepted or rejected: For each scenario, it records whether $d_t$ was accepted, and the decision is made by selecting the action (accept/reject) that received the majority of votes. This voting mechanism allows the agent to leverage information from multiple simulated trajectories to make more robust decisions. 

\subsubsection{Our Method (DQN)}\label{sec:ours}

We use a neural network model as an agent for the decision layer, trained via reinforcement learning. The algorithm solves the MDP formulation of the problem using the observation vector defined in the DQVRP model section. Specifically, we adopt a \textit{Double DQN}~\cite{doubleq} algorithm, which employs both a policy network and a target network to improve convergence speed and stability. At each step, the model computes a \emph{vehicle assignment}, deciding whether to accept a destination and which vehicle to assign it to. We refer to this method as \texttt{DQN-VA}. Assigning a demand to a vehicle before knowing the full scenario, though overly restrictive, stabilizes routing and thus aids learning.

In this DQN method, a neural network approximates the Q-value function to select the best action. Since the action space is finite, the network takes an observation as input and outputs the Q-values for all actions. Using an $\epsilon$-greedy exploration strategy, the action with the highest Q-value is chosen with probability $1-\epsilon$, and a random action is selected with probability $\epsilon$. The network is trained to approximate the optimal Q-value function by updating its parameters based on received rewards.

We use an exponential $\epsilon$-decay schedule: $\epsilon$ starts at 1 and decreases exponentially to a minimum value $\epsilon_\text{limit}$. This allows the agent to gradually shift from exploration to exploitation, balancing the exploration-exploitation trade-off.

Target network parameters are updated using a hybrid strategy combining hard and soft updates. Hard updates fully replace the target network parameters with the policy network parameters at fixed intervals, while soft updates gradually blend a fraction of the policy network parameters into the target network. We perform a hard update every $N$ steps, with soft updates applied in between using a mixing coefficient $\eta$. This approach maintains stability while allowing the target network to follow important changes in the policy network. 

The model is a 3-layer MLP with 1024 neurons per layer. Training uses the AdamW optimizer~\cite{adamw} on a train set of 500 scenarios, while evaluation is conducted on 100 test scenarios.

\section{Experiments} \label{sec:experiments}

We conducted experiments on both real supply chain operations and synthetic instances using 100 test scenarios each. We focus on the fully dynamic case (DoD = 1) to assess the ability of our methods to make strategic decisions under uncertainty. Performance across other DoD values is detailed in Section~\ref{appx:dods}.

Two types of synthetic instances are considered. The \textit{uniform} instance draws destinations uniformly across a plane, while the \textit{clustered} instance draws four random centers and samples destinations from a normal distribution around each center. For the realistic instance, possible destinations and their probability distribution $P$ were extracted from a transporter's operational data. In synthetic instances, all demands have equal probability.

\begin{figure}[hbt!]
    \centering
    \includegraphics[width=0.75\linewidth]{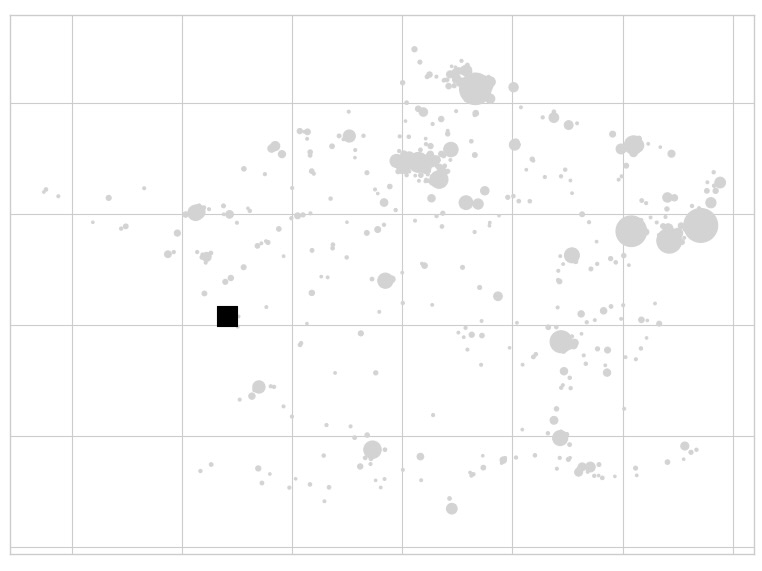}
    \includegraphics[width=0.75\linewidth]{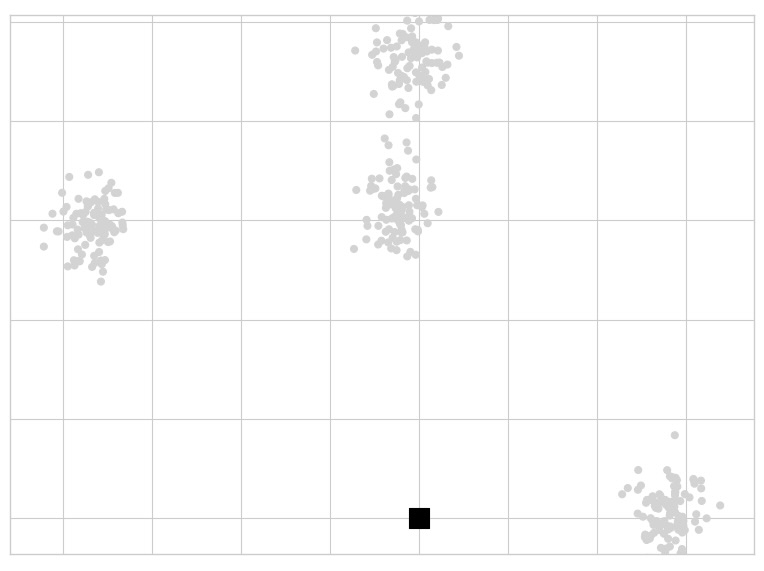}
    \includegraphics[width=0.75\linewidth]{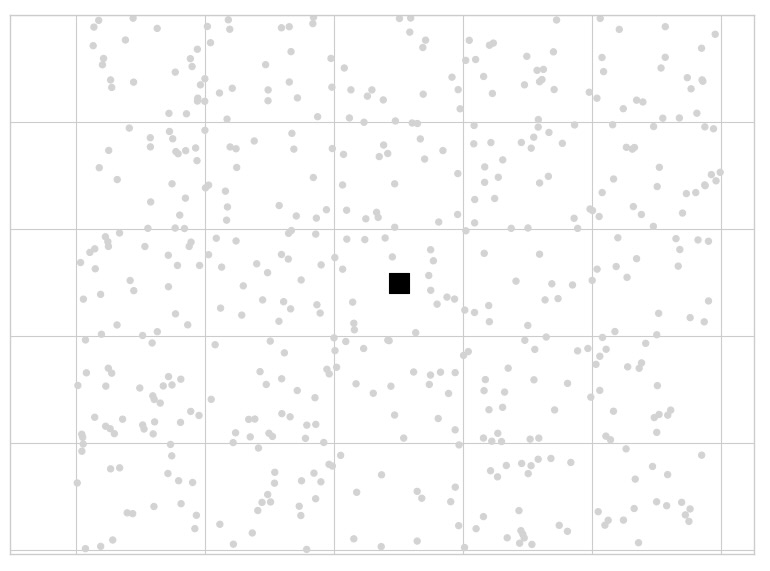}
    \caption{Destinations of the realistic instance (top), clustered instance (middle), and uniform instance (bottom).}
    \label{fig:data}
\end{figure}

The uniform distribution evaluates agent performance in scenarios with minimal spatial structure and high uncertainty, reflecting cases such as e-commerce. The clustered distribution introduces spatial patterns that agents must exploit, creating a more challenging setting. The spatial distributions of the three instance types are visualized in Figure~\ref{fig:data}.

Experiments were run on an M1 MacBook Pro, except for the \texttt{MSA} algorithm, which required a 60-core computing node. Training a DQN model takes approximately 3 hours per instance type; execution times are reported in Table~\ref{tab:algorithm_times}.

Further details on algorithm parameters, instances, and solutions produced are provided in the Appendix. All code, experiments, data, and models are open source and will be released after the anonymous review.

\begin{table}[ht] 
\centering 
\begin{tabular}{lccc} \hline 
Algorithm & Realistic & Clustered & Uniform \\
\hline 
\texttt{FAFS} & $3.25\pm0.11$ & $0.75\pm0.79$ & $3.05\pm0.11$ \\ 
\texttt{DQN-VA} & $0.13\pm0.01$ & $0.08\pm0.12$ & $0.18\pm0.08$ \\ 
\texttt{MSA} & $473\pm4$ & $97\pm1$ & $363\pm3$ \\ \hline \end{tabular}
\caption{Average execution times (in seconds) to solve a scenario from the three types of instances. Average computed over 100 samples with a 95\% confidence interval.
\label{tab:algorithm_times} } 
\end{table}

\subsection{Performance Comparison for Different Instance Types}

\begin{figure*}[ht!]
    \centering
    \includegraphics[width=0.3\linewidth]{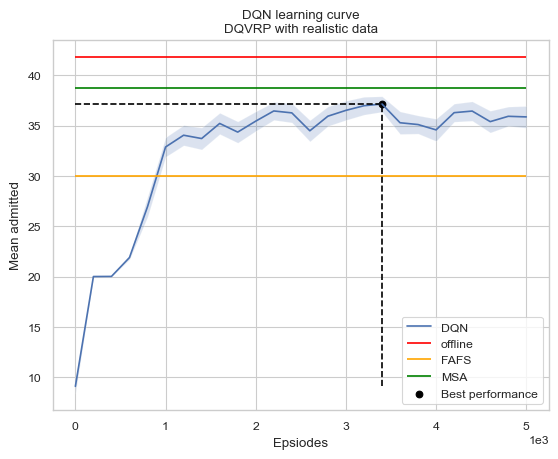}
    \includegraphics[width=0.3\linewidth]{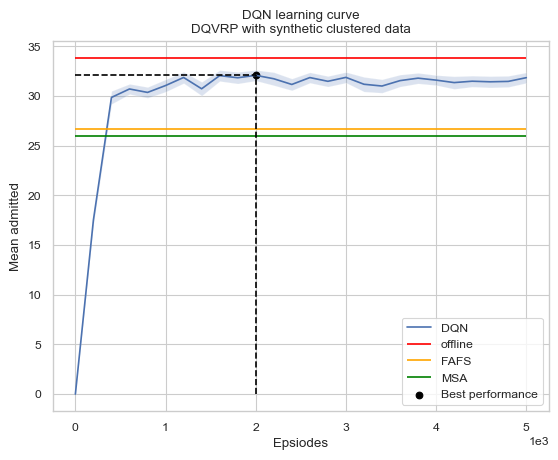}
    \includegraphics[width=0.3\linewidth]{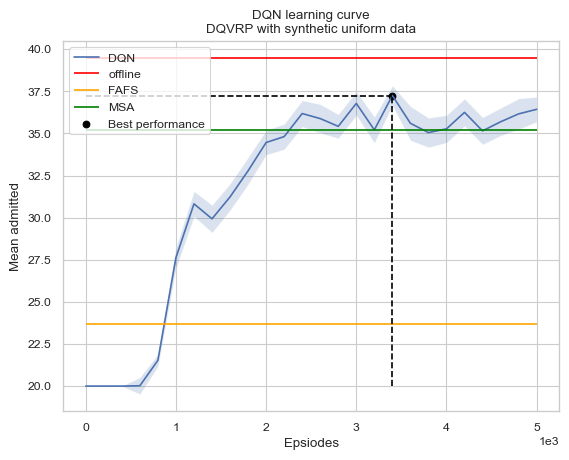}
    \caption{Learning curves on the three datasets with DoD~=~1. The shaded areas indicate 95\% confidence intervals. The horizontal lines mark the mean performance $\Ad$ of the baseline methods.}
    \label{fig:rl}
\end{figure*}

\begin{figure}[ht!]
    \centering
    \includegraphics[width=0.67\linewidth]{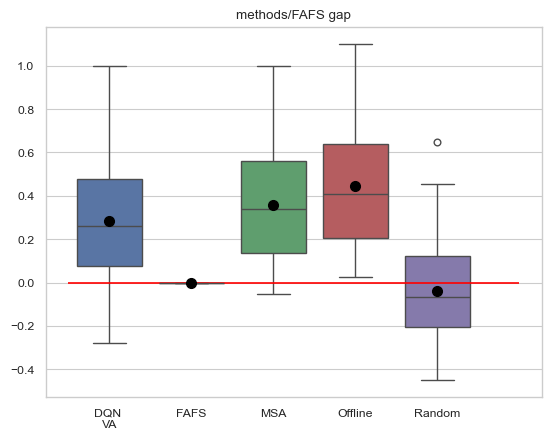}
    \caption{Distribution of performance improvements over \texttt{FAFS} on the realistic instance scenarios (means shown as points).}
    \label{fig:improvement-real}
\end{figure}

We begin by analyzing the results obtained on scenarios of the realistic instance. Figure~\ref{fig:rl} illustrates both the learning process of the RL model and the mean accepted demands achieved by each method. The comparative performance analysis is presented in Figure~\ref{fig:improvement-real}, which shows the improvement achieved over the baseline \texttt{FAFS} method.

\texttt{DQN-VA} demonstrates significant performance improvements over \texttt{FAFS} in terms of admitted demand. As Figure~\ref{fig:improvement-real} demonstrates, this performance is achieved consistently across the majority of test scenarios, rather than being explained by a few large improvements in pathological cases.
While \texttt{DQN-VA} achieves marginally lower performance compared to \texttt{MSA}, it offers a significant computational advantage. Specifically, \texttt{MSA} requires 101 simulations per decision step, leading to prohibitively high computational overhead.

\begin{figure}[ht!]
    \centering
    \includegraphics[width=0.67\linewidth]{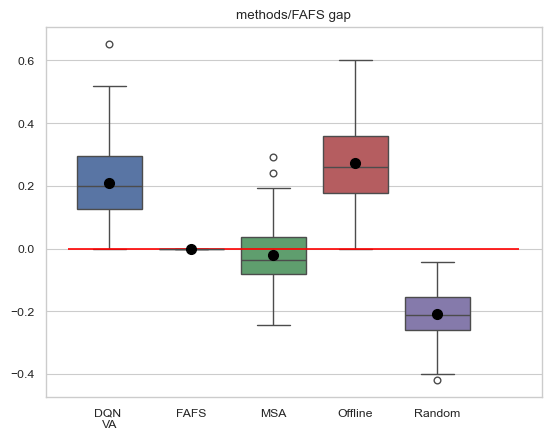}
    \caption{Distribution of performance improvements over \texttt{FAFS} on the clustered instance scenarios (means shown as points).}
    \label{fig:improvement-cluster}
\end{figure}

In the clustered dataset, the \texttt{DQN-VA} approach outperforms the \texttt{MSA} method, achieving performance levels very close to the \texttt{Offline} baseline, as demonstrated in Figure~\ref{fig:improvement-cluster}. Notably, \texttt{MSA} performs even worse than the simple \texttt{FAFS} baseline. This poor performance can be attributed to the uniform request probability distribution across destinations, combined with the specific spatial distribution of destinations, which creates challenging conditions for sampling-based algorithms.

\begin{figure}[ht!]
  \centering
    \includegraphics[width=0.67\linewidth]{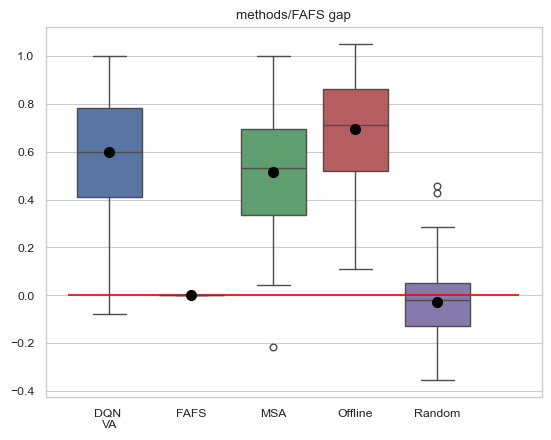}
  \caption{Distribution of performance improvements over \texttt{FAFS} on the uniform instance scenarios (means shown as points).}
  \label{fig:improvement-uniform}
\end{figure}

Furthermore, in the uniform dataset, the \texttt{DQN-VA} approach once again outperforms the \texttt{MSA} method and is considerably better than other online methods, as demonstrated in Figure~\ref{fig:improvement-uniform}. 
These results further demonstrate the dependence of sampling-based methods on the probability distribution of requests across destinations. In contrast, our RL method shows resilience even in challenging scenarios with uniform spatial distribution of destinations and limited knowledge of future demands. This robustness highlights a key advantage of our approach.

\subsection{Sensitivity To An Uncertain Horizon}

We evaluate the robustness of our RL method to observation noise, focusing on uncertainty in the planning horizon $H$, which represents the total number of future delivery requests. In practice, $H$ is rarely known exactly and must be estimated, potentially with significant error. To assess the impact of this uncertainty, we test our method under different levels of estimation noise by adding Gaussian perturbations of 10\%, 20\%, and 30\% standard deviation to the true value of $H$. We also consider a setting where no estimate of $H$ is provided, which serves as a lower bound. For each noise level, a separate model is trained on scenarios from the realistic instance.

\begin{figure}[ht!]
    \centering
    \includegraphics[width=0.85\linewidth]{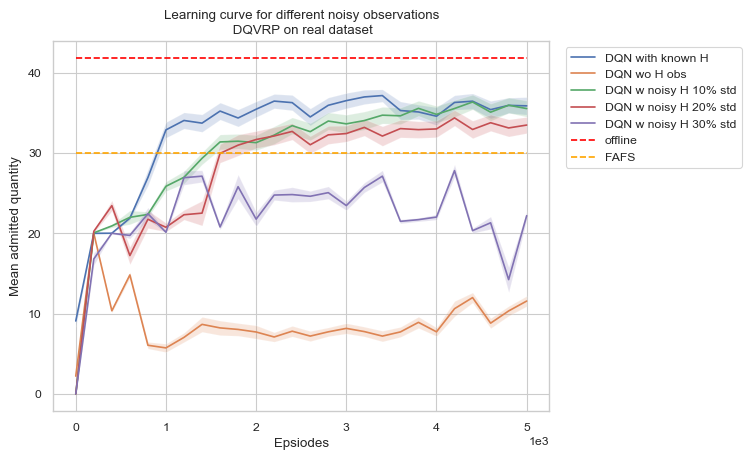}
    \caption{Learning curves of the models with different observations for the $H$
value.}
    \label{fig:obs variation}
\end{figure}

As shown in Figure~\ref{fig:obs variation}, the model trained without any information about H performs substantially worse, indicating that horizon information is crucial for \texttt{DQN-VA}. Nonetheless, the method exhibits graceful degradation as noise increases. With horizon estimates corrupted by up to 20\% standard deviation, \texttt{DQN-VA} continues to outperform \texttt{FAFS}. This robustness suggests that the approach remains effective in practical settings where the planning horizon is uncertain but can be reasonably estimated.

\subsection{Impact of the DoD}\label{appx:dods}

We evaluate the performance of \texttt{DQN-VA} across different Degrees of Dynamism (DoD) on the realistic instance. The same model trained with DoD = 1 is used for all experiments. We run 100 identical test scenarios while varying the DoD to quantify the impact of dynamism.
\begin{figure}[ht!]
    \centering
    \includegraphics[width=0.65\linewidth]{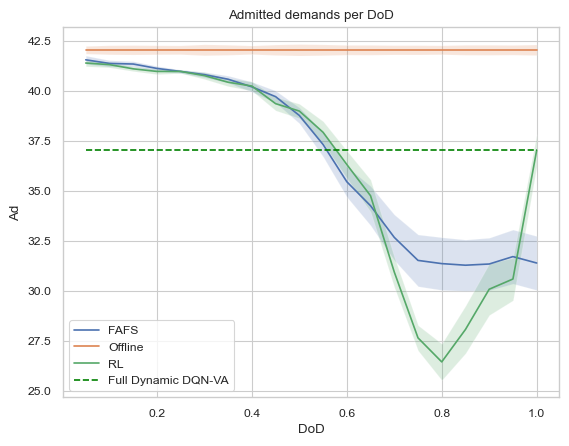}
    \caption{Admitted demands (mean with 95\% confidence interval)  of \texttt{DQN-VA}, \texttt{FAFS} and \texttt{Offline} baseline across different values of the DoD.}
    \label{fig:dods}
\end{figure}

As shown in Figure~\ref{fig:dods}, performance of \texttt{FAFS} deteriorates as dynamism increases, while the performance of \texttt{Offline} remains constant as it ignores the dynamicity of the problem. \texttt{DQN-VA} performs best for DoD values close to 1 and becomes comparable to \texttt{FAFS} for DoD~$< 0.7$. This behavior reflects two factors: (i) the model is trained under fully dynamic conditions, and (ii) when DoD~$< 0.7$, most decisions occur in the static phase, reducing the relative impact of dynamic decision-making.

A limitation of \texttt{DQN-VA} is its lower performance for DoD in $[0.6, 1)$ compared to DoD = 1, despite having access to more information. This is because \texttt{DQN-VA} fixes vehicle assignments during the static phase—typically saturating low-emission vehicles—and cannot revise these assignments during the dynamic phase, unlike \texttt{FAFS}. This issue can be mitigated by running \texttt{DQN-VA} as if the setting were fully dynamic whenever DoD~$> 0.6$, which consistently yields much better performance than \texttt{FAFS}.

\section{Discussion}

We introduced an MDP-based approach to the dynamic CVRP with an emission quota and stochastic requests (DQVRP), with the objective of maximizing the number of served destinations. Our results show that reinforcement learning can substantially improve decision quality by enabling anticipatory and strategic accept/reject choices. Leveraging engineered state features, the proposed model learns to discard incompatible requests and assign feasible ones to suitable vehicles.

A key strength of \texttt{DQN-VA} is its invariance to the planning horizon $H$. For a fixed fleet configuration, a single trained model can operate on instances with different horizons without retraining, and it maintains strong performance even when $H$ is only approximately known. This property enhances scalability and supports deployment in environments where planning horizons vary or are estimated with uncertainty.

This work can be extended to settings with heterogeneous customer demands. Preliminary results, provided in the Appendix, indicate that additional feature design and algorithmic adjustments are required to achieve competitive performance in this variant.

More broadly, the two-layer structure of our framework is compatible with diverse dynamic VRP variants, including pickup-and-delivery and same-day delivery problems, as well as settings with heterogeneous vehicle capacities. The assignment layer can also be combined with more powerful routing algorithms, benefiting from incremental re-optimization as new requests appear.

Future research could explore advanced neural architectures, such as transformers, to better model the sequential structure of the assignment procedure. Their ability to capture global dependencies may enhance the representation of interactions between vehicles, accepted requests, and pending demands, thereby improving the sequential decision-making formulation of the static QVRP and its dynamic extensions.

\section*{Acknowledgment}
This work was supported by the French LabCom HYPHES (ANRS-21-KCV1-0002).

\newpage
\bibliography{refs}

\newpage

\appendix

\section{Details on the Instances and the Algorithms}\label{appx: exp details}

\subsection{Construction of the Instances}
\begin{figure}[hbt!]
    \centering
    \includegraphics[width=.7\linewidth]{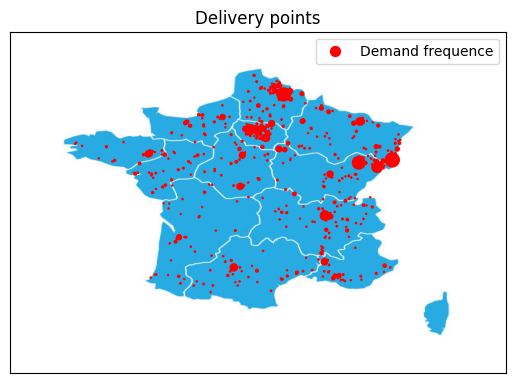}
    \caption{Demands and their frequencies of the realistic data of the deliveries in France on the map.}
    \label{annex:map}
\end{figure}

We rigorously evaluated and compared our methods through extensive experiments using both realistic and synthetic datasets across diverse test scenarios.
Our experimental design aims to assess the methods' effectiveness under varying degrees of complexity that mirror real-world logistics challenges. The realistic data was collected from actual delivery operations (see Figure~\ref{annex:map}), providing authentic patterns and constraints. To complement this, we generated synthetic datasets specifically designed to stress-test our methods from multiple perspectives - including clustered and uniform spatial distributions. This comprehensive approach allows us to evaluate both the practical applicability and robustness of our methods compared to classical ones. The spatial distribution characteristics of these datasets are visualized in Figure \ref{fig:data}, highlighting the distinct patterns that each dataset presents to the optimization algorithms.

The realistic dataset represents a typical supply chain scenario, where client locations exhibit both clustering in specific areas and dispersion across others. The dataset also captures varying delivery request frequencies among clients, reflecting real-world demand patterns.

The clustered dataset simulates a scenario with clients concentrated in distinct geographical zones, with substantial distances between clusters.
The uniform dataset presents a scenario where clients are evenly distributed across the service area. This configuration is particularly valuable for analysis as it represents situations where transporters must operate with minimal predictive information about future demands, testing the robustness of decision-making strategies under high uncertainty.

Each dataset has its specifities regarding the vehicles, the number of possible destinations, the distribution of these destinations, and the a priori probability of request from each destination. In this situation, the agents are expected to effectively leverage these specifities to develop and implement strategic approaches for their assignment.
For the realistic data, we consider a fleet of 4 vehicles with a quota of 50 chosen to be challenging. Meanwhile, for the synthetic data we consider a fleet of 2 vehicles subject to a quota of 100. In both cases, the fleet has two types of vehicles : hybrid and diesel, such that half of the fleet is of the first type and the other of the second. We have $H=100$ for the real and uniform dataset, and $H=50$ for the clustered dataset.

We generate 100 test instances for each scenario. For the realistic data, the frequency of the deliveries provided by the dataset are used as a probability vector to draw destinations. While for the synthetic datasets this probability is uniformly distributed.
The distance matrices are computed using the euclidean distance.

Our instances utilize four vehicles with different emission factors: two hybrid (0.15), and two diesel (0.3).
All vehicles have the same capacity equals to 20, which is chosen with respect to the number of vehicles and the total capacity necessary for delivering all packages.

\subsection{Hyperparameters}
\begin{table}[htbp]
    \centering
    \caption{Hyperparameters.}
    \begin{tabular}{ p{4cm} p{2.5cm} }
      \toprule
        \textbf{Hyperparameters}           & \textbf{Values}                  \\
      \toprule
      \textbf{Simulated Annealing for the routing layer}&\\
      \midrule%
        $\tau_{\text{init}}$ & 1 000           \\
        $\tau_{\text{limit}}$   & 1        \\
        Cooling rate     & 0.995   \\
        Max iterations     & 50 000   \\
      \midrule%
      \midrule%
      
      \textbf{RL Agent (DQN)}&\\
      \midrule%
        Layers for $\leq 2$ vehicles       & $[512,\ 512,\ 512]$         \\
        Layers for $> 2$ vehicles        & $[1024,\ 1024,\ 1024]$        \\
        Activation function                    & ReLU     \\
        $\gamma$                    & 0.99        \\
        $\epsilon_\text{limit}$                    & 0.05        \\
        $\epsilon$ decay period                   & 1 000 episodes       \\
        Hard target update interval $N$                    & 100        \\
        Soft target update parameter $\eta$                    & 0.01        \\
        Learning rate                   & 1e-4        \\
        Mini batch size                   & 128        \\
        $k_\text{AD}$                   & 3        \\
        $k_\text{FD}$                   & 7        \\
      
      

        
            \bottomrule
        \end{tabular} 
\end{table}

\section{Routes Generated by Different Methods}
\label{appx:routing}
\begin{figure}[htb!]
    \centering
    \begin{subfigure}[b]{0.53\linewidth}
        \includegraphics[width=\linewidth]{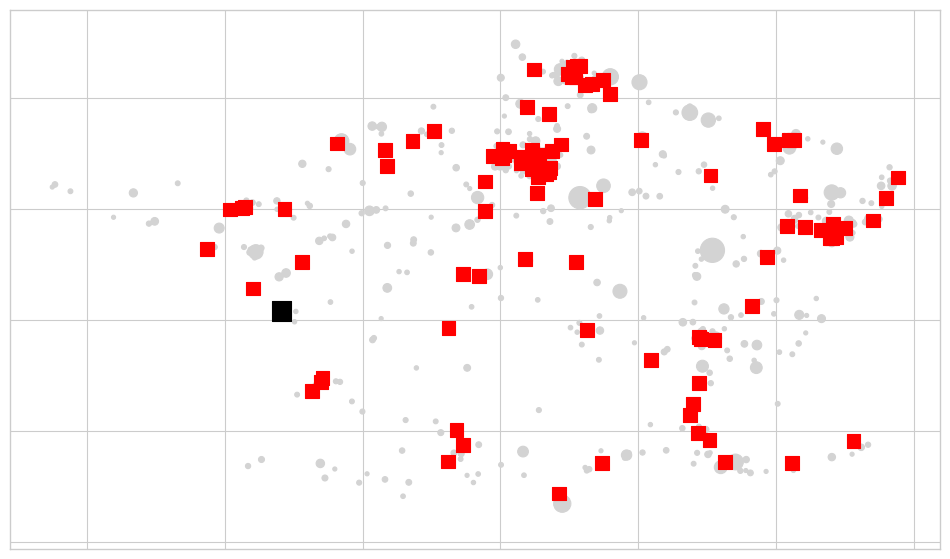}
        \caption{The destinations.}
    \end{subfigure}
    \begin{subfigure}[b]{0.67\linewidth}
        \includegraphics[width=\linewidth]{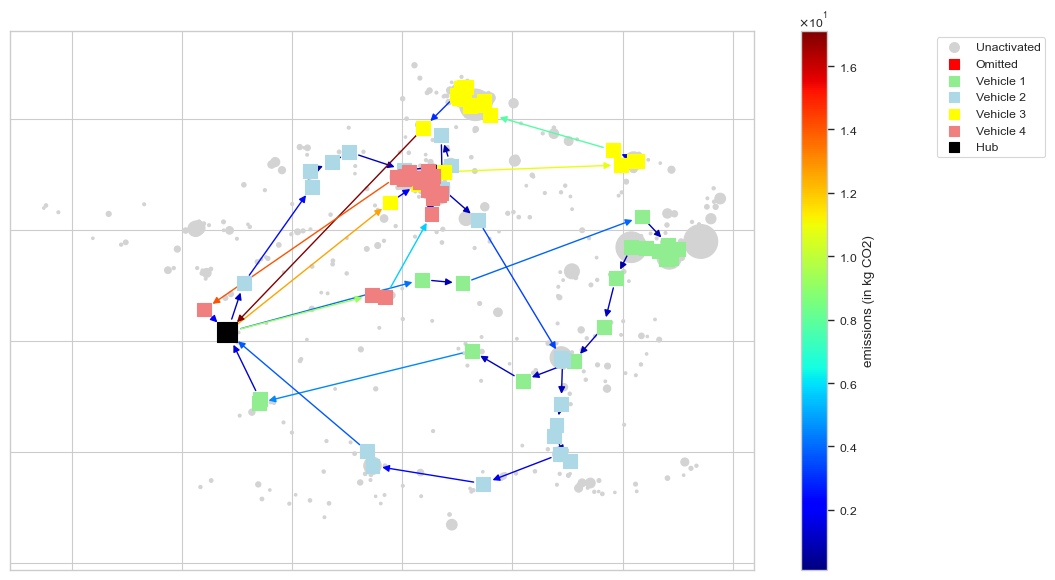}
        \caption{Routes of \texttt{Offline} without emission quota.}
    \end{subfigure}
    \hfill
    \begin{subfigure}[b]{0.67\linewidth}
        \includegraphics[width=\linewidth]{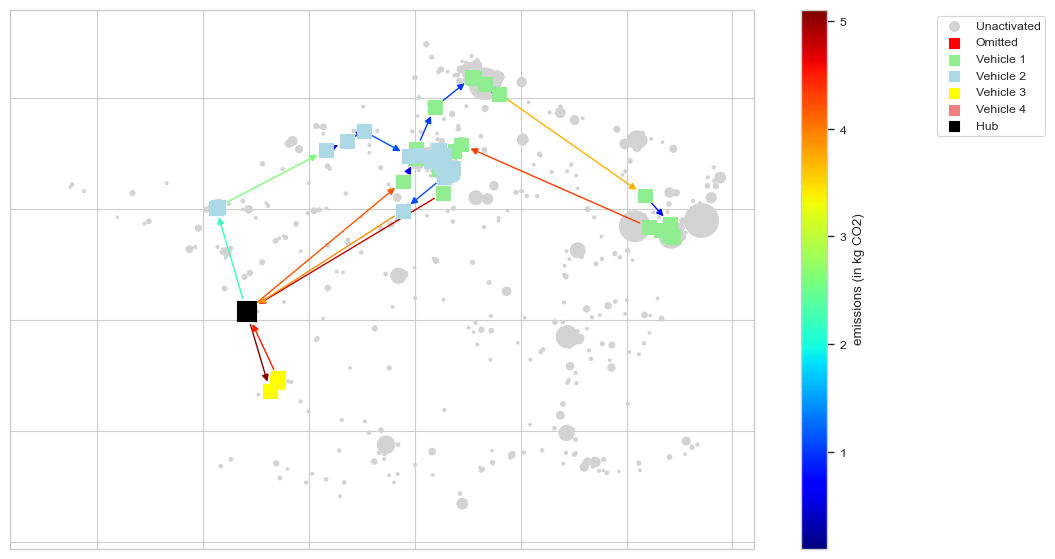}
        \caption{\texttt{Offline}}
    \end{subfigure}
    \begin{subfigure}[b]{0.67\linewidth}
        \includegraphics[width=\linewidth]{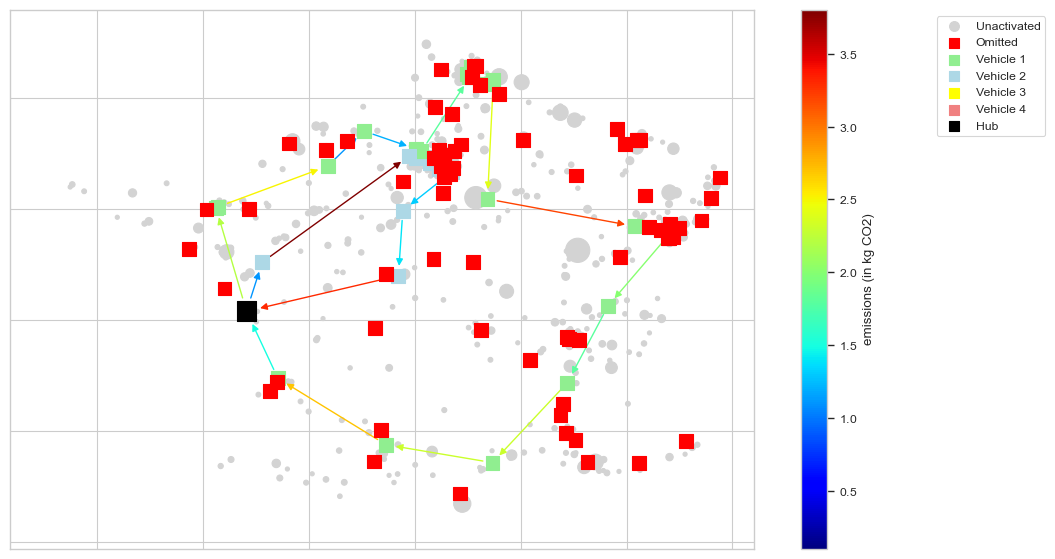}
        \caption{\texttt{FAFS}}
    \end{subfigure}
    \hfill
    \begin{subfigure}[b]{0.67\linewidth}
        \includegraphics[width=\linewidth]{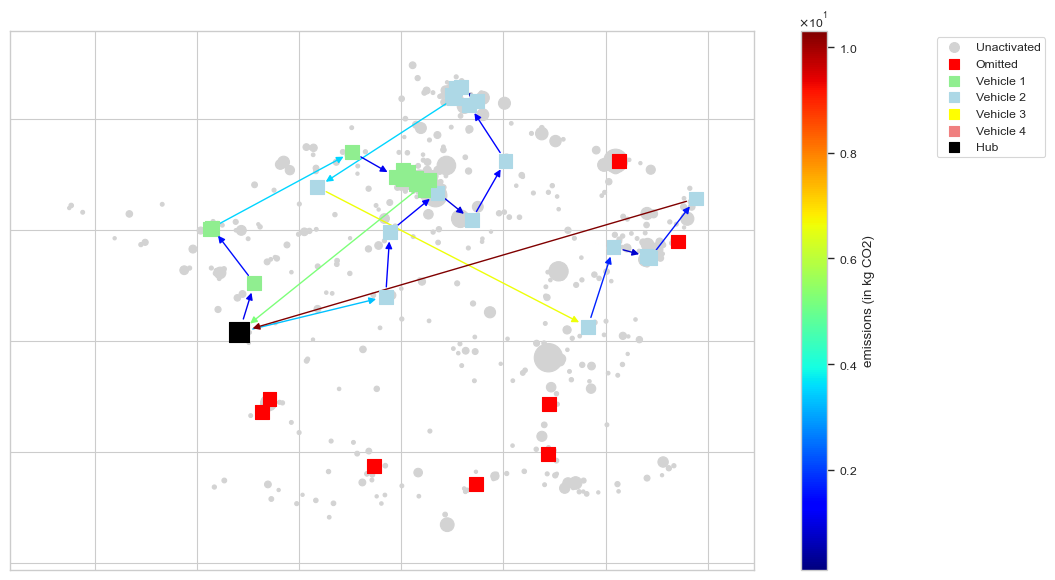}
        \caption{\texttt{DQN-VA}}
    \end{subfigure}
    \begin{subfigure}[b]{0.67\linewidth}
        \includegraphics[width=\linewidth]{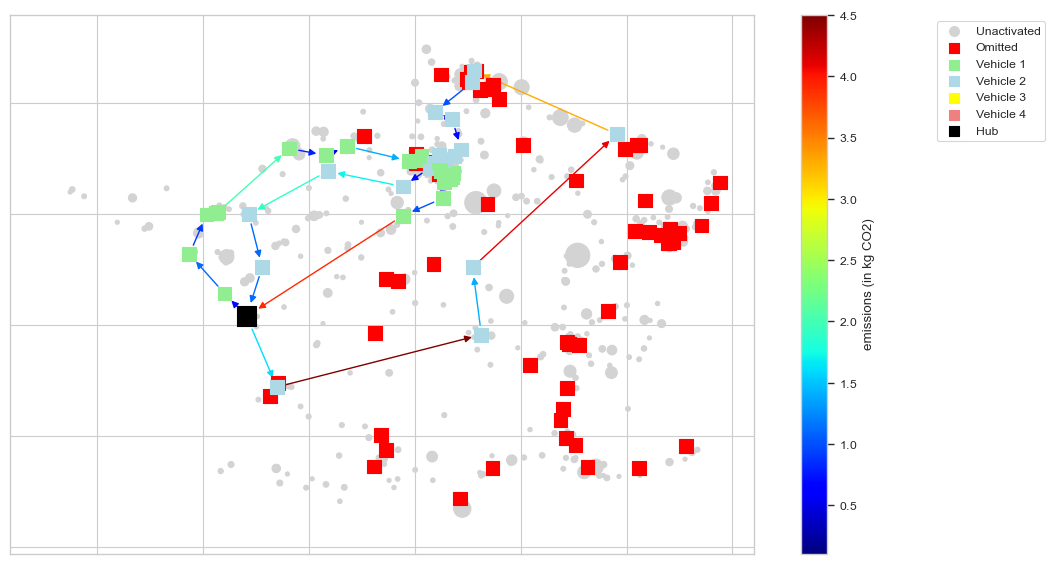}
        \caption{\texttt{MSA}}
    \end{subfigure}
    \hfill
    \caption{Routes generated by different methods for realistic data.}
    \label{fig:routes-real}
\end{figure}

\begin{figure}[htb!]
    \centering
    \begin{subfigure}[b]{0.75\linewidth}
        \centering
        \includegraphics[width=\linewidth]{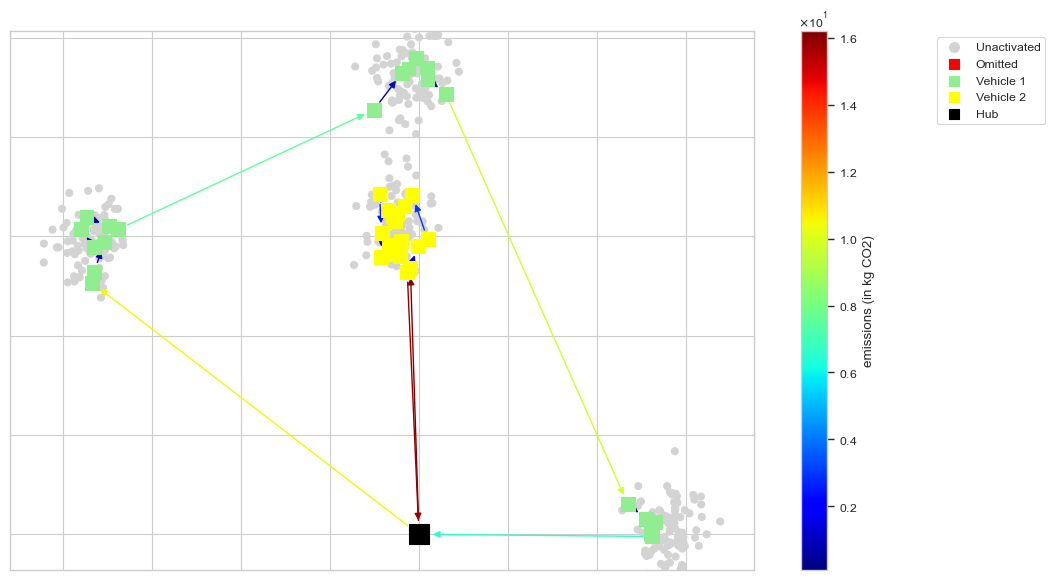}
        \caption{\texttt{Offline}}
    \end{subfigure}
    \hfill
    \begin{subfigure}[b]{0.75\linewidth}
        \centering
        \includegraphics[width=\linewidth]{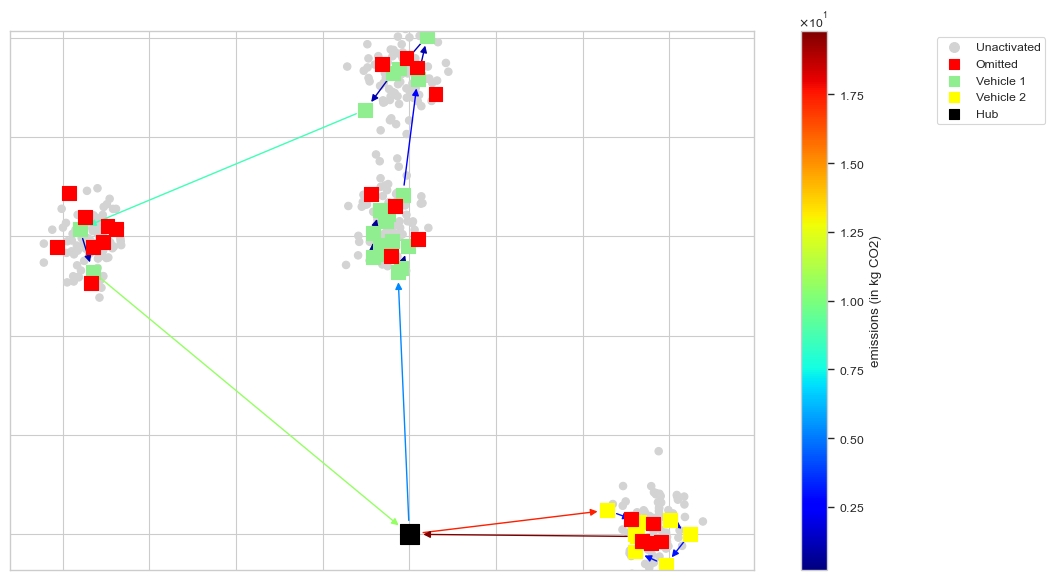}
        \caption{\texttt{FAFS}}
    \end{subfigure}
    \hfill
    \begin{subfigure}[b]{0.75\linewidth}
        \centering
        \includegraphics[width=\linewidth]{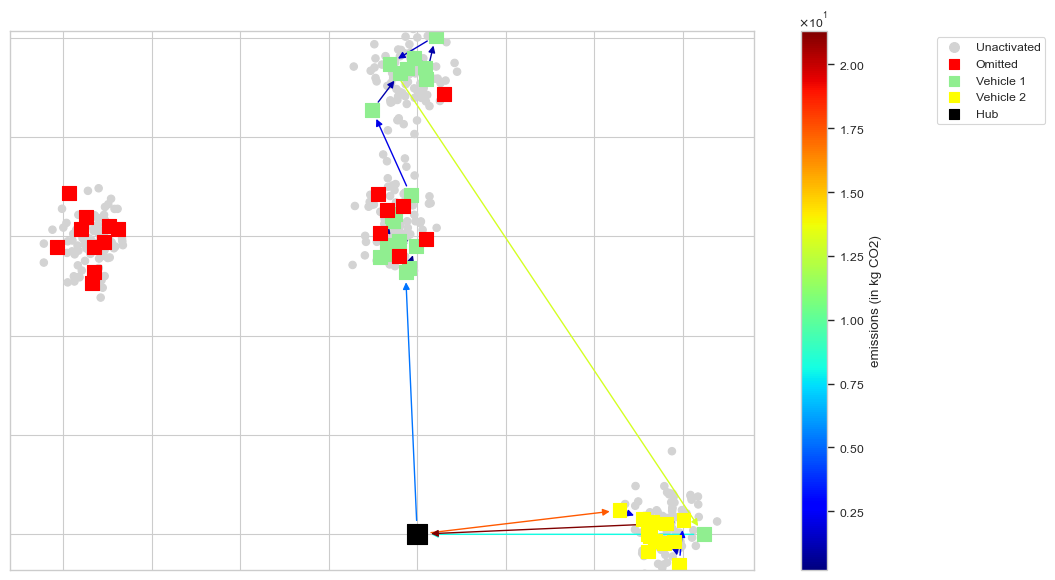}
        \caption{\texttt{DQN-VA}}
    \end{subfigure}
    \begin{subfigure}[b]{0.75\linewidth}
        \centering
        \includegraphics[width=\linewidth]{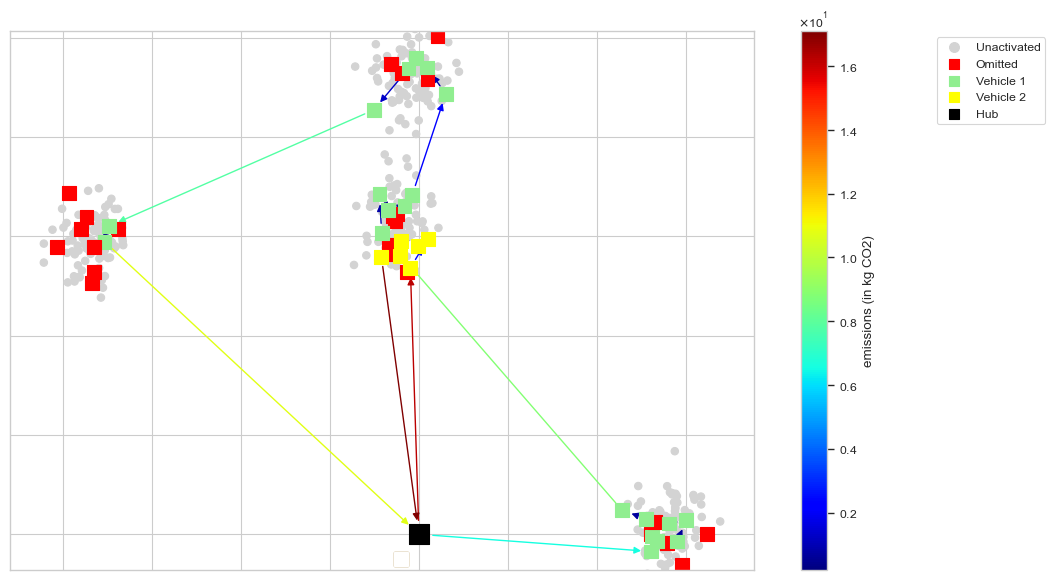}
        \caption{\texttt{MSA}}
    \end{subfigure}
    \caption{Routes generated by different methods for clustered data.}
    \label{fig:routes-clusters}
\end{figure}

\begin{figure}[hbt!]
    \centering
    \begin{subfigure}[b]{0.75\linewidth}
        \centering
        \includegraphics[width=\linewidth]{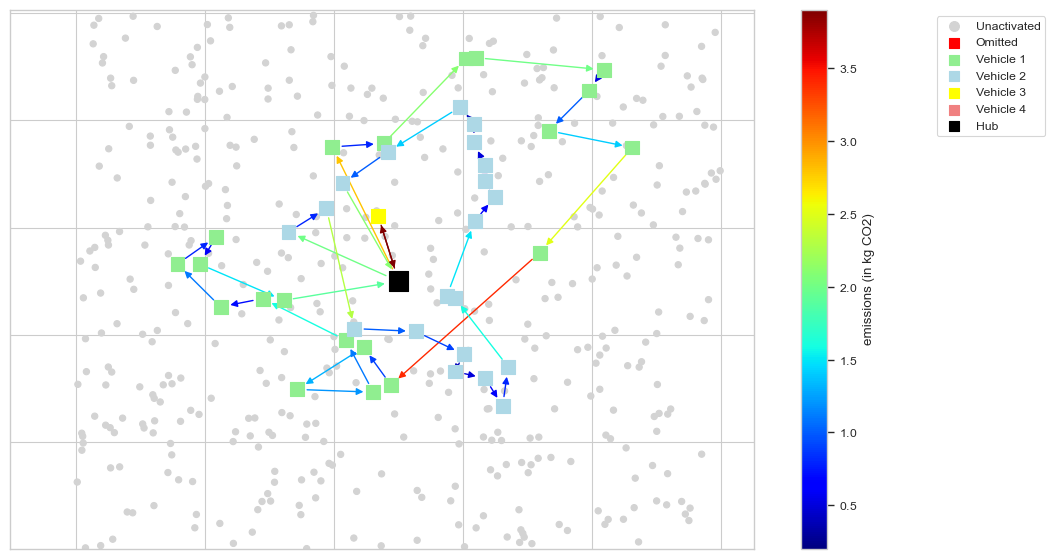}
        \caption{\texttt{Offline}}
    \end{subfigure}
    \begin{subfigure}[b]{0.75\linewidth}
        \centering
        \includegraphics[width=\linewidth]{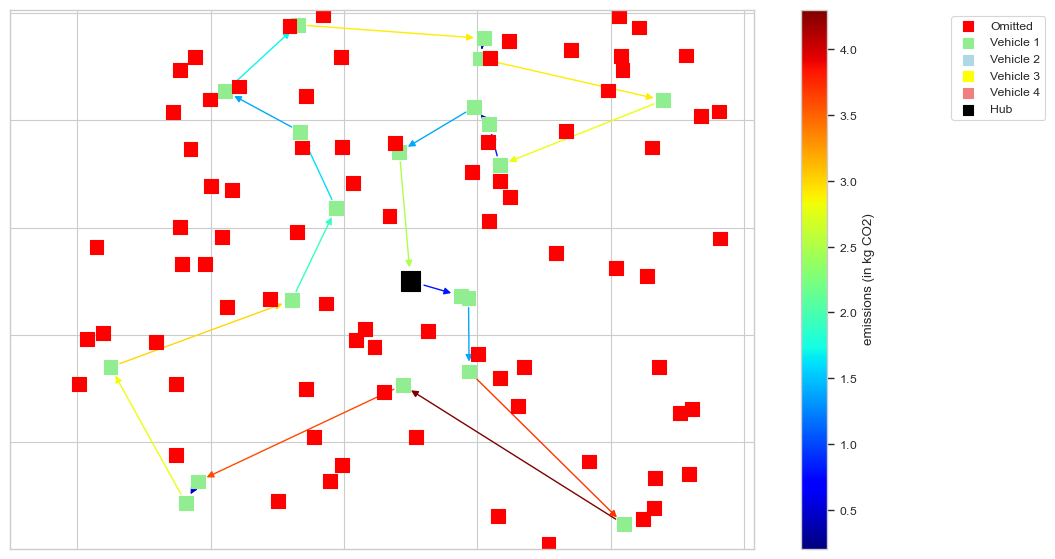}
        \caption{\texttt{FAFS}}
    \end{subfigure}
    \hfill
    \begin{subfigure}[b]{0.75\linewidth}
        \centering
        \includegraphics[width=\linewidth]{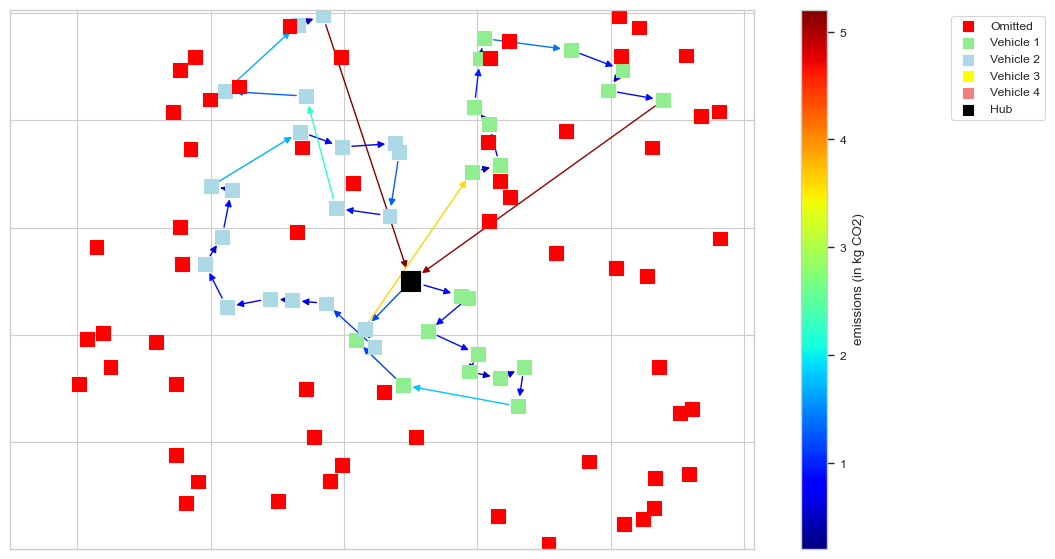}
        \caption{\texttt{DQN-VA}}
    \end{subfigure}
    \begin{subfigure}[b]{0.75\linewidth}
        \centering
        \includegraphics[width=\linewidth]{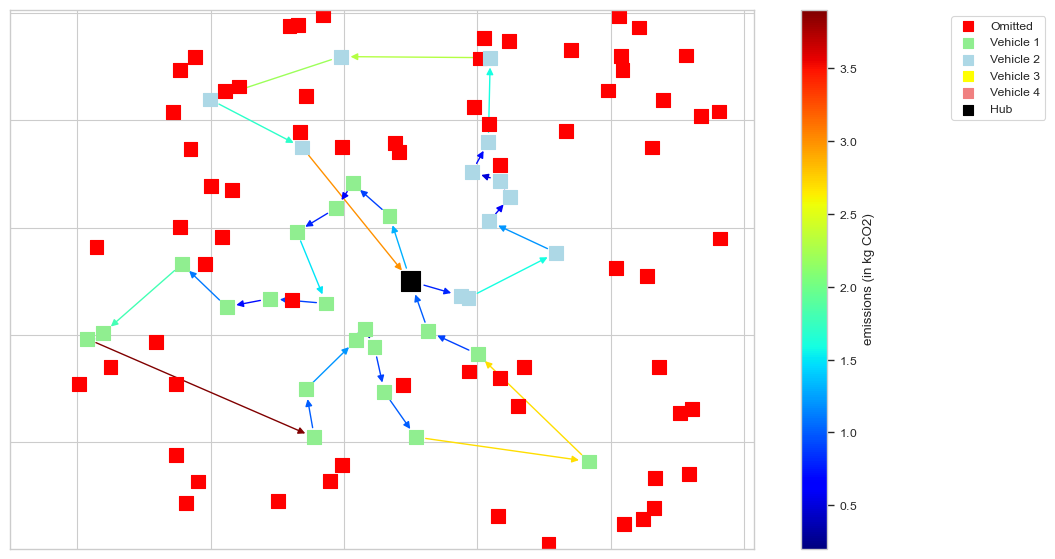}
        \caption{\texttt{MSA}}
    \end{subfigure}
    \hfill
    \caption{Routes generated by different methods for uniform data.}
    \label{fig:routes-uniform}
\end{figure}

Figures~\ref{fig:routes-real}, \ref{fig:routes-uniform} and \ref{fig:routes-clusters} illustrate the routes generated by each method in different datasets, providing a visual comparison of their respective routing. Each method demonstrates distinct patterns in how they construct routes due to their assignment strategy.

\section{Heterogeneous Quantities}

To further evaluate our methods' performance with more realistic scenarios, we conducted additional experiments where destinations have heterogeneous demand quantities, in contrast to our previous experiments where all destinations had uniform unit demands.

These experiments used the realistic dataset comprising 50 destinations and a heterogeneous fleet of 4 vehicles (2 hybrid and 2 diesel vehicles). The demand quantities were generated according to the following formula:
$q = \mathbf{1}_H + ((C-1)*M - H/2) * \lfloor{X}\rfloor$
where $C$ represents the vehicle capacity, $M$ is the number of vehicles, and $X$ follows a Dirichlet distribution with parameter vector $\mathbf{\alpha} = \mathbf{1}_H$. To ensure feasibility, we regenerated $X$ whenever the total demanded quantity exceeded the total fleet capacity.
The training process for the RL methods follows the same approach outlined in previous sections, but with the reward at time $t$ that is $q_t$. Moreover, we add the quantity demanded to the observations.

\begin{figure}[ht!]
    \centering
    \includegraphics[width=0.65\linewidth]{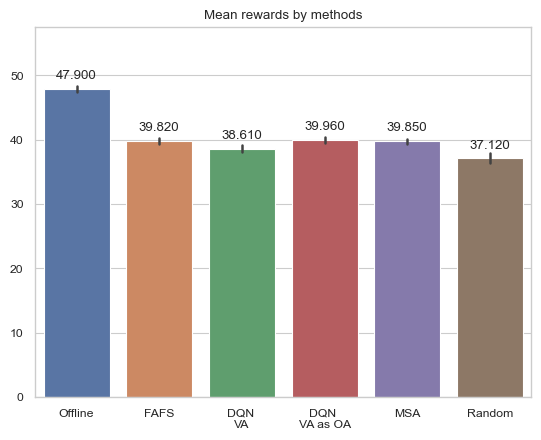}
    \caption{Comparison of total admitted quantities across different methods on realistic dataset with heterogeneous demand quantities.}
    \label{fig:dqrvp heter qs}
\end{figure}

With the DoD fixed at 1, we evaluate the performance of \texttt{DQN-VA} against benchmark methods in this more complex setting. 
Moreover, we test another version of a DQN model called \texttt{DQN-VAasOA} where we use the same model as \texttt{DQN-VA}, but the vehicle assignment is determined by the routing layer and we accept the demand if the model chooses a non-zero action. This model allows us to change the vehicle assignment of other demands when a new demand is added. This helps to avoid rejecting demands because of the capacity constraint.

As demonstrated in Figure~\ref{fig:dqrvp heter qs}, \texttt{DQN-VA} had lower performance compared to \texttt{FAFS}, \texttt{MSA} and \texttt{Offline} baseline. However, \texttt{DQN-VAasOA} outperforms slightly all the online methods.

This outcome reveals a limitation of our method when dealing with non-unit quantities. We hypothesize that the primary challenge comes from the observation space containing limited information about future demand quantities, which prevents the agent from developing and learning meaningful strategies in this environment. 
The combined uncertainty regarding both future destinations and demanded capacity quantities has a significant impact on overall performance.
These results may vary based on the distribution of demanded quantities. Further research is needed to fully understand how different demand distributions impact the overall performance.
This suggests that additional features or architectural modifications may be necessary to better handle heterogeneous demand quantities. Further, to help the agent, it is possible to integrate the action masks to remove the vehicles that cannot accept the incoming demand.

\end{document}